\documentclass[conference]{IEEEtran}
\IEEEoverridecommandlockouts
\usepackage{cite}  
\usepackage{tabu}                      
\usepackage{booktabs}                  
\usepackage{lipsum}                    
\usepackage{mwe}                       
\usepackage{epsfig}
\usepackage{subcaption}
\usepackage{calc}

\usepackage{amstext}
\usepackage{amsmath}
\usepackage{amssymb}
\usepackage{amsthm}
\usepackage{multicol}
\usepackage{multirow}
\usepackage{soul}
\usepackage{bm}
\usepackage{pgf}
\usepackage{tikz}
\usetikzlibrary{spy}
\usepackage{tikzscale}
\usetikzlibrary{arrows,external}
\usetikzlibrary{spy}
\usepackage{pgfplots}
\usepackage[skins]{tcolorbox}
\usepackage{import}
\usepackage{url}
\usepackage{caption}
\usepackage[ruled,vlined]{algorithm2e}

\usepackage{algorithmic}
\usepackage{makecell}
\usepackage[table]{xcolor}
\definecolor{colorTrd}{rgb}{0.95,0.95,0.65}
\definecolor{colorSnd}{rgb}{1, 0.85, 0.7}
\definecolor{colorFst}{rgb}{1, 0.7, 0.7}

\usepackage{threeparttable}

\usepackage{adjustbox}
\usepackage{rotating}
\def\BibTeX{{\rm B\kern-.05em{\sc i\kern-.025em b}\kern-.08em
    T\kern-.1667em\lower.7ex\hbox{E}\kern-.125emX}}
\begin{document}
\title{Sketch2Motion: Text-driven 2D Sketch to 3D Animation via
Diffusion-guided Skeleton Optimization}
\author{
\IEEEauthorblockN{Gaurav Rai\IEEEauthorrefmark{2}, 
Ojaswa Sharma\IEEEauthorrefmark{2}}

\IEEEauthorblockA{\IEEEauthorrefmark{2}Graphics Research Group, IIIT Delhi\\
gauravr@iiitd.ac.in, ojaswa@iiitd.ac.in}
}

\maketitle
\begin{abstract}
Animation of 2D hand-drawn sketches provides an effective medium for visual communication. However, these sketches pose challenges, particularly in handling occlusions and accurately mapping motion. While 3D animation naturally addresses these challenges, estimating 3D motion remains a very complex task. Recent approaches to converting 2D sketches to 3D animations have mainly focused on specific types of motion, such as bipedal movements and facial expressions. We propose Sketch2Motion, a diffusion-guided framework for skeleton-based motion synthesis that combines classical character animation pipelines with deep generative priors. Our method represents motion using skeletal transformations, which are propagated to mesh deformations via linear blend skinning. To guide the resulting animation toward realistic and semantically meaningful motion, we integrate a text-to-video diffusion model via motion-aware score-distillation sampling (MoSDS), enabling optimization without paired motion data.
Additionally, we apply physics-inspired smoothness, topological, and contact constraints to stabilize optimization and preserve motion plausibility. Further, we integrate a spring-mass simulator to introduce secondary motion effects. The proposed framework is generalized, fully differentiable, modular, and compatible with biped, quadruped, and non-living articulated characters. Experiments demonstrate that our approach produces temporally coherent, text-aligned animations that outperform baseline motion transfer methods that lack generative priors or explicit physical constraints. We will make our code and dataset publicly available.
\end{abstract}

\begin{figure*}
    \includegraphics[width=0.9\linewidth]{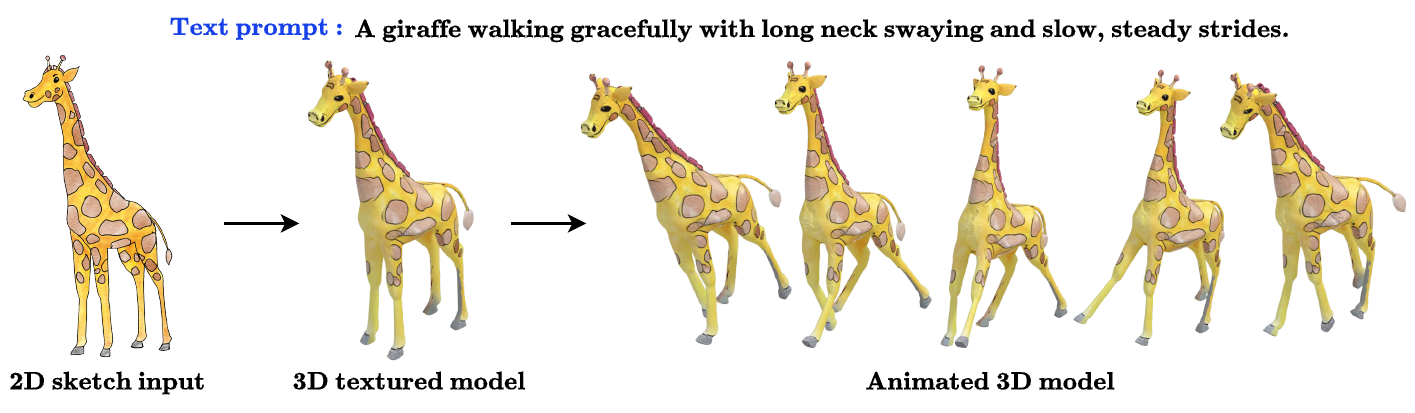}
    \centering
    \caption{Overview of our proposed method for 2D sketch to 3D animation.}
\label{fig:teaser}
\end{figure*}

\section{Introduction}
Animating 2D hand-drawn sketches poses technical challenges arising from the inherent ambiguity and sparsity of 2D representations. Unlike photographs or 3D scans, sketches lack explicit depth cues, surface geometry, and consistent topology, making it difficult to infer plausible 3D structure and motion. The primary challenge lies in estimating and simulating complex 3D motion from inherently 2D inputs, which requires bridging a significant dimensional gap while preserving the artistic intent, stylization, and abstraction characteristic of hand-drawn content. Minor errors in depth inference or articulation can easily lead to visually implausible or artistically inconsistent animations. Another major challenge arises from occlusions or self-intersections commonly present in sketches, where overlapping strokes may represent different body parts or structural components. Such occlusions introduce ambiguity in part ordering and connectivity, complicating both motion estimation and temporal coherence during animation. Additionally, sketches often exhibit varying levels of detail and incomplete contours, further increasing uncertainty in part segmentation and motion interpretation. As a result, many existing sketch animation methods~\cite{su2018live, rai2024sketchanim} are constrained to simple motions or rely on strong assumptions about object structure and pose.

Traditional approaches to sketch animation have therefore been limited to specific motion categories~\cite{smith2023method} or predefined templates~\cite{patel2016tracemove, gupta2018sheetanim}, restricting their applicability to a broader range of characters and motion styles. To address these limitations, recent methods for 2D sketch-to-3D animation~\cite{feng2017magictoon, weng2019photo, zhou2024drawingspinup, yoon2025occlusion} maintain the texture style of drawing but rely on template-based motion for animation. Animating 3D characters with realistic motion remains a challenging problem. Classical motion capture pipelines rely on expensive hardware setups and carefully curated datasets. Moreover, adapting captured motion to new characters requires non-trivial retargeting procedures. Learning-based motion synthesis methods offer a promising alternative. However, they typically depend on large-scale paired datasets for training. In addition, these methods often struggle to generalize across varying skeletal topologies and non-standard character shapes. These limitations are particularly apparent when animating sketch-derived or stylized 3D models, for which large-scale motion datasets are unavailable. Recent advances in diffusion models have demonstrated impressive performance in image and video generation, suggesting their potential as strong priors for motion synthesis~\cite{poole2022dreamfusion}. Text-to-video diffusion models, in particular, encode rich semantic and temporal information about motion patterns. However, directly applying diffusion models to 3D animation is non-trivial due to the high dimensionality of articulated motion spaces, the lack of paired supervision between text and skeletal motion, and the need to preserve physical plausibility and temporal topological consistency. 

Recent, Text-to-3D model animation methods lack physics-based constraints~\cite{wu2025animateanymesh, wang2026bimotion} and soft-body dynamics~\cite{li2025articulated, mu2025smp} and are limited to a specific kind of motion~\cite{guo2025make}. To address these limitations, our framework begins by inflating the input 2D sketch into a 3D representation~\cite{li2025step1x}, and generates a 3D mesh that preserves the visual appearance and stylistic characteristics of the original sketch. In this work, we bridge classical animation pipelines and modern generative modeling by introducing a diffusion-guided, skeleton-based motion synthesis framework. We utilize a pretrained text-to-video diffusion model~\cite{wang2023modelscope} as a high-level semantic prior while explicitly optimizing skeletal motion parameters. We enhance score distillation sampling (SDS)~\cite{poole2022dreamfusion} with motion-aware SDS (MoSDS), extracting gradients from the diffusion model to guide motion optimization toward text-aligned, visually plausible animations without updating the diffusion model itself. To ensure stability and realism, we incorporate physics-inspired regularization terms, including temporal smoothness, topological joint limits, symmetry constraints, and ground contact penalties. We integrate a spring-mass simulator to add secondary motion effects that make a character feel alive. Our framework is fully differentiable, does not require paired motion supervision, and generalizes across a wide range of character types, including bipeds, quadrupeds, flying, aquatic, and non-living objects. Together, these components provide a principled, extensible solution for controllable, realistic skeleton-driven motion synthesis as illustrated in Figure~\ref{fig:teaser}. Our contributions are as follows:
\begin{itemize}
    \item A framework that applies motion-aware score distillation from a text-to-video diffusion model to directly optimize skeletal motion parameters, enabling semantic motion synthesis without paired data while generalizing across diverse articulated entities.
    \item We introduce a novel Kinematic non-uniform rational basis spline conditioning framework for MoSDS. 
    \item A set of physics-inspired constraints, including temporal smoothness, joint's range of motion, and ground contact, that are essential for stable and realistic diffusion-guided motion optimization.
    \item A spring–mass simulation framework designed to estimate and apply secondary motion effects to object components, along with their associated elements.
    \item We provide a dataset of 100 hand-drawn sketches spanning diverse object categories and structural complexities.
\end{itemize}

\section{Related work}

\subsection{Text-driven sketch Animation}
Text-driven sketch animation has recently emerged as an active research area, utilizing the generative priors of large text-to-video diffusion models. Early work by Gal et al.~\cite{gal2024breathing} demonstrates that pretrained text-to-video priors can animate static sketches without task-specific training data, showing the generalization capability of diffusion models beyond natural images. Their approach utilizes score distillation sampling (SDS)~\cite{poole2022dreamfusion} to optimize sketch parameters by distilling gradients from a pretrained diffusion model, effectively bridging the gap between static abstraction and temporal dynamics. Building on this principle, ESA~\cite{rai2024enhancing} highlights key challenges in diffusion-guided sketch animation, including temporal flickering and shape distortion, and introduces rigidity-awareness via an as-rigid-as-possible (ARAP)~\cite{igarashi2005rigid} loss and temporal consistency constraints, with length-area regularization to stabilize the optimization. Concurrently, several works explore structured representations for sketch animation. AniClipart~\cite{wu2025aniclipart} parameterizes clipart motion using keypoint skeleton and uses cubic B\'ezier motion trajectory for each joint, combining video diffusion guidance with ARAP regularization to preserve object rigidity. It is further extended by pushing the boundaries of deformation quality. FlexiClip~\cite{khandelwal2025flexiclip} incorporates temporal jacobians and flow-matching losses to mitigate the geometric distortions common in pure SDS approaches. Similarly, Dynamic Typography~\cite{liu2025dynamic} adapts this diffusion-guided strategy to typographic animation, employing neural displacement fields and vector-based constraints to ensure that semantic motion does not compromise letter legibility. In the realm of storytelling, FairyGen~\cite{zheng2025fairygen} integrates Multi-Modal LLMs for storyboard generation with cinematic shot design, enabling consistent, narrative-driven animation of character sketches.

Parallel efforts have explored raster-based techniques and advanced generative control to handle complex appearances and motions. AnimateSketches~\cite{deng2025animatesketches} introduces instance-aware masks and attention-based guidance to animate multi-object raster sketches, reducing interference between moving elements. Similarly, FlipSketch~\cite{bandyopadhyay2025flipsketch} further stabilizes raster animation through a reference-frame mechanism and DDIM~\cite{song2020denoising} inversion, allowing for expressive "flip-book" style motion that preserves the sketch's artistic identity. More recent work extends text-driven sketch animation toward richer motion semantics and multi-object scenarios. Beyond single-subject animation, works by Liu et al.~\cite{liu2025multi} and Liang et al.~\cite{liang2025multi} move beyond single-object animation by decomposing scenes into multiple sketch entities and performing compositional motion planning guided by diffusion priors and large language models. Despite this progress, existing methods primarily rely on 2D stroke deformation, raster warping, or keypoint-based regularization and lack explicit kinematic structure, limiting physical plausibility and controllability. In contrast, these limitations motivate approaches that integrate diffusion guidance with explicit skeletal representations and physics-aware constraints to enable structured, stable, and semantically meaningful 2D sketches to 3D model animations.

\subsection{2D Sketch to 3D animation}
Translating 2D sketches into 3D animated models presents a fundamental challenge due to the inherent depth ambiguity and abstraction of drawn strokes. Early approaches to 2D sketch-to-3D animation focus on inflating static drawings into simple 3D representations using heuristic or interactive modeling. MagicToon~\cite{feng2017magictoon} pioneers interactive 2D-to-3D generation by introducing a mobile augmented reality (AR) system. It automatically converts 2D cartoon drawings on paper into textured 3D models, allowing users to effortlessly animate and interact with their creations directly in real-world AR environments. Advancing this concept to real-world human subjects, Photo Wake-Up~\cite{weng2019photo} animates characters from single photographs. Rather than relying on simple 2.5D puppets, it warps a 3D parametric body template (SMPL) to match the subject's 2D silhouette. It further projects normal and skinning maps to lift the figure into a fully rigged 3D mesh capable of complex, out-of-frame 3D animations. Emphasizing highly accessible, from-scratch creation, Monster Mash~\cite{dvorovzvnak2020monster} proposes a single-view approach to casual 3D modeling and animation. By combining sketch-based 3D inflation with a novel rigidity-preserving, layered deformation model, the method transforms 2D outlines into smooth 3D meshes. These meshes can be interactively animated through a simple click-and-drag interface. This approach does not require explicit skeletal rigging or multi-view part merging.
While these approaches enable intuitive sketch-based animation, they rely heavily on user interaction, predefined motion patterns, or simplified geometry, and thus struggle to produce complex, realistic motion or generalize across diverse character types.

Recent work has increasingly shifted toward learning-based, automated pipelines that infer 3D structure, rigging, and motion from 2D sketches. These approaches have significantly advanced the translation of static hand-drawn inputs into dynamic animations by addressing domain gaps and underlying structural complexities. Sketch2Anim~\cite{zhong2025sketch2anim} tackles the direct conversion of sketch storyboards into 3D animations by using a neural mapper that aligns 2D sketches with 3D keyposes and trajectories, acting as a direct control mechanism for a 3D conditional motion generator. To preserve the unique stylistic identity of the original artwork, Smith et al.~\cite{smith2025animating} introduce an approachable 2.5D character model and a view-dependent retargeting technique that successfully applies complex 3D skeletal motions, including transverse rotations, to single childlike figure drawings. Concurrent works have targeted specific bottlenecks in stylization and secondary dynamics during motion synthesis. Addressing the loss of style and contour flickering caused by overlapping body parts, Yoon et al.~\cite{yoon2025occlusion} propose an occlusion-robust stylization framework that leverages optical flow to provide consistent edge guidance during dynamic 3D motions. Meanwhile, Zhou et al.~\cite{zhou2025rigging} focus on the natural animation of complex non-rigid elements, such as flowing hair and cloth, and introduce a hybrid system that grounds the character using traditional skeletal animation for geometric consistency, while injecting secondary dynamics via video diffusion models to achieve expressive, realistic motion.

Despite significant progress, most existing methods either require specialized training data, a domain gap between abstract sketches and realistic 3D priors, assume specific sketch styles, or decouple 3D reconstruction from motion semantics, motivating the need for approaches that can integrate the generative flexibility of diffusion models with explicit structural constraints to achieve robust, physically plausible 3D animation from 2D sketches.

\subsection{3D animation}
3D animation has recently seen a paradigm shift driven by the integration of large-scale diffusion priors and unified representation learning. While traditional methods relied heavily on manual rigging or expensive motion capture data. A significant challenge in 3D animation is the lack of high-quality 3D motion data compared to the abundance of 2D video. Consequently, recent works focus on distilling motion priors from pre-trained video diffusion models to drive 3D synthesis.
Xie et al.~\cite{xie2025animamimic} introduce AnimaMimic, which animates static 3D meshes by automatically synthesizing a monocular video to construct a skeleton and skinning weights. It then refines the animation using differentiable rendering and physically grounded soft-tissue simulation. Similarly, Li et al.~\cite{li2025articulated} present Articulated Kinematics Distillation (AKD), which integrates video priors with structural controls. By applying Score Distillation Sampling (SDS) directly to a skeleton-based representation, AKD drastically reduces the degrees of freedom to joint-level controls, distilling complex motions while preserving structural consistency. Expanding this control to arbitrary topologies, Animax~\cite{huang2025animax} bridges video priors with skeleton-based animation by representing 3D motion as multi-view, multi-frame 2D pose maps. Using a joint video-pose diffusion model, it triangulates these sequences into 3D joints and applies inverse kinematics to animate diverse meshes. To tackle category-free pose transfer across distinct character types, Chai et al.~\cite{chai2025mimicat} propose MimiCAT, a cascade-transformer approach that uses semantic keypoints to establish flexible, soft many-to-many correspondences rather than relying on strict one-to-one mappings. Addressing 4D synthesis from single monocular videos, Chen et al.~\cite{chen2026motion} introduce Motion 3-to-4. This method adopts a feed-forward framework that decomposes the task into static 3D shape generation and motion reconstruction. It uses a frame-wise transformer to predict per-frame vertex trajectories over a canonical reference mesh, resulting in temporally coherent geometry. Recently, Wang et al.~\cite{wang2026bimotion} propose BiMotion, a novel approach for text-guided 3D character animation that represents motion as continuous B-spline curves rather than discrete frame sequences. This formulation enables smoother, more coherent, and variable-length motions while improving alignment with textual descriptions.

Text-to-4D generation aims to synthesize dynamic 3D scenes directly from textual descriptions. Recent advancements have moved towards foundation models that generalize across diverse mesh topologies.
In this direction, Wu et al.~\cite{wu2025animateanymesh} introduce AnimateAnyMesh, a feed-forward 4D foundation model. Unlike optimization-based methods that require per-subject tuning, this model learns a universal motion representation applicable to any mesh topology, significantly reducing inference time. To address motion quality, Su et al.~\cite{sun2025animus3d} introduce Animus3D, which uses Motion Score Distillation (MSD). This technique improves upon standard Score Distillation Sampling (SDS) by specifically targeting motion dynamics, resulting in smoother and more coherent animations driven by text prompts. To further enhance temporal stability, Sun et al.~\cite{sun2025tracking} introduce a tracking-guided 4D generation framework. By integrating foundation-model-based point trackers into the generation process, it constrains the motion priors to follow physically consistent trajectories, reducing the drift often seen in generative 4D. Similarly, Rahamim et al.~\cite{rahamim2024bringing} tackle 4D generation from static objects, focusing on a training-free approach. It utilizes view-consistent noise injection to hallucinate motion, enabling the animation of 3D objects without extensive retraining or fine-tuning.

While generative approaches focus on surface deformation, other works aim to innovate the underlying structural representations of animation, specifically, rigging, skinning, and physics-based control. RigMo~\cite{zhang2026rigmo} presents a unified framework that jointly learns rig parameters and motion latents. This joint learning approach ensures that the generated rigs are explicitly optimized for the intended motions, overcoming the limitations of decoupled pipelines that treat rigging and animation as separate tasks. In this direction, Zhang et al.~\cite{zhang2026skin} introduced Skin Tokens, a compact, learned representation for autoregressive rigging. By tokenizing skinning weights, they enable the use of transformer-based architectures to predict complex deformation properties efficiently. In the domain of character posing, Guo et al.~\cite{guo2025make} present Make-It-Poseable, a feed-forward model that predicts latent posing parameters for 3D humanoid characters. This method simplifies the traditional rigging process, allowing for rapid character articulation. Finally, integrating physics into the generative pipeline, \cite{mu2025smp} introduces Score-Matching Motion Priors (SMP). This work focuses on physics-based character control, learning reusable motion priors that can satisfy physical constraints, thereby bridging the gap between kinematic generation and dynamic simulation.
In our method, we overcome the limitation of physics-based constraints by introducing physics-based regularization and secondary motion to generate natural motion.

\section{Methodology}

\begin{figure*}
    \centering
    \includegraphics[width=0.95\linewidth]{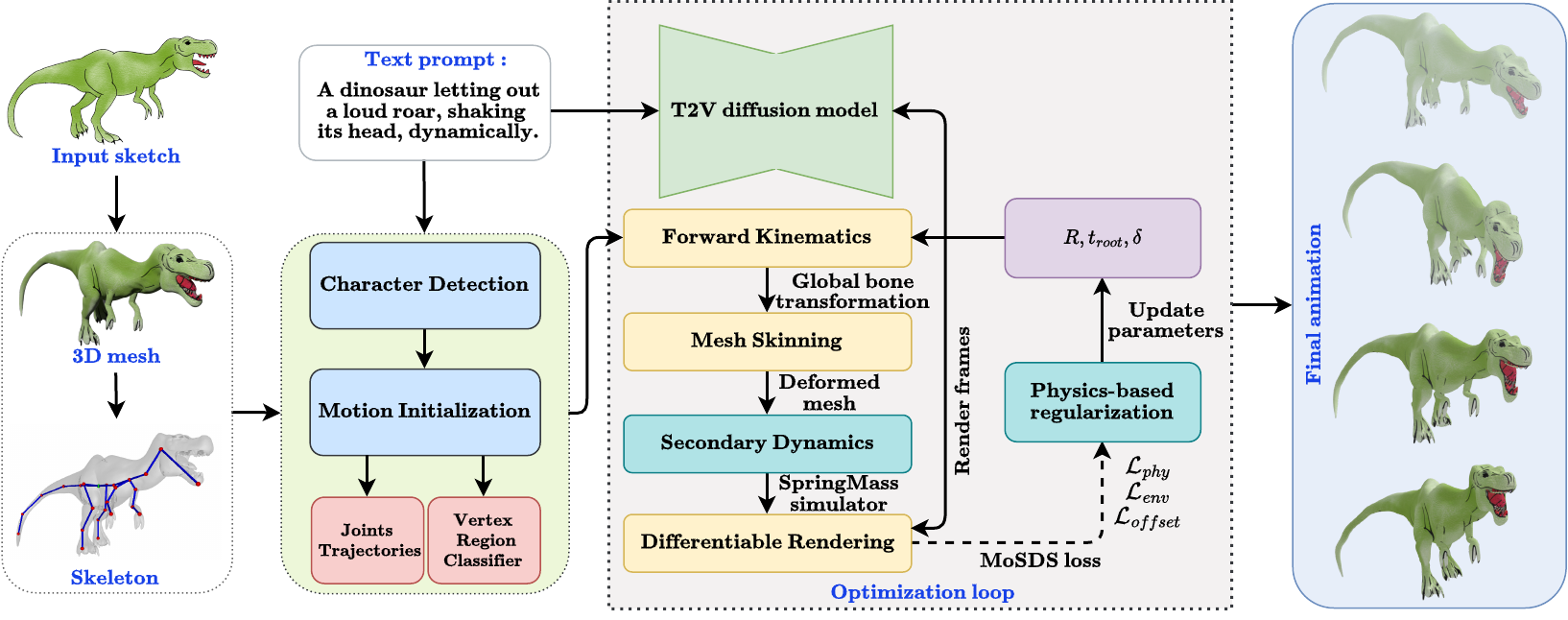}
    \caption{Overview of pipeline architecture of our proposed method for 2D sketch to 3D model animation.}
    \label{fig:pipleine}
\end{figure*}

\subsection{Overview}
We use a 2D sketch $S$ as input and first inflate it into a textured 3D mesh model. Thereafter, we automatically infer a kinematic skeleton and joint hierarchy from the 3D geometry. The extracted skeleton defines joint locations, parent-child relationships, and a rest pose. Initially, we perform motion initialization to identify the object class and determine an appropriate motion type, enabling the assignment of a valid range of motion for each joint type in the skeleton.
To generate animation, we condition a text-to-video (T2V) diffusion model on the textured mesh and the extracted skeleton. We optimize the skeletal parameters in the latent space of a pre-trained T2V diffusion model~\cite{wang2023modelscope} guided by a user-specified text prompt describing the desired motion. We use motion-aware score distillation sampling (MoSDS) to ensure semantically aligned, meaningful, and temporally coherent motion generation. By parameterizing the shape with a kinematic skeleton, we constrain the model to generate physically plausible articulated poses that preserve the object's structural rigidity while enabling flexible and controlled motion. We use the T2V model not as a direct video generator, but as a differentiable semantic prior. Given the text prompt $y$, we optimize the joint rotations over time such that the rendered sequence of the rigged mesh aligns with the learned motion priors of the diffusion model. This optimization-by-synthesis approach allows us to generate complex, non-linear motions that are semantically faithful to the text while remaining kinematically grounded in the 3D structure. 
Further, to generate physically plausible motion, we use physics-based regularization and a spring-mass simulator to model secondary motion effects. Figure~\ref{fig:pipleine} illustrates the pipeline of the proposed method. Algorithm~\ref{alg:sketchanim_updated} provides a set of steps of our proposed method.

\subsection{Skeleton representation}
\label{sec:skel_represent}
We represent an articulated object using a kinematic skeleton composed of $J$ joints organized in a hierarchical tree structure with a root joint. A joint $j$ is defined by a rest-pose position and a parent index $p(j)$. To facilitate efficient kinematic computations, we pre-compute the rest-pose offsets (bone vectors) $\mathbf{o}_j = \mathbf{x}_j^{\text{rest}} - \mathbf{x}_{p(j)}^{\text{rest}}$ for all non-root joints, while the root joint's offset is defined relative to the global origin, where $\mathbf{x}$ denotes a joint's coordinate. The topological order of the hierarchy is computed via depth-first search to ensure that parent transformations are always processed before their children during the forward pass. 

To generate dynamic, expressive animations, we parameterize motion over a sequence of $T$ frames using three distinct sets of learnable variables. First, the global root translation $\mathbf{t}_{\text{root}} \in \mathbb{R}^{T \times 3}$ controls the character's global trajectory and positioning in world space. Second, joint rotations $\mathbf{R} \in \mathbb{R}^{T \times J \times 3}$ are represented in axis-angle form (rotation represented as $\theta\hat{\mathbf{n}}$ with angle of rotation $\theta \in \mathbb{R}$ and direction unit-vector $\hat{\mathbf{n}}\in \mathbb{R}^3$) and ensure smooth gradient flow. These define the relative orientation of each joint with respect to its parent. Finally, we introduce local offsets $\boldsymbol{\delta} \in \mathbb{R}^{T \times J \times 3}$, a per-joint translational residual that models non-rigid secondary dynamics that cannot be captured by only rigid skeletal rotation. Given the rest-pose offset $\mathbf{o}_j$ of joint $j$ relative to its parent, we define a local rigid transformation $\mathcal{T}(\mathbf{R}_{t,j}, \mathbf{o}_j) \in \mathbb{R}^{4 \times 4}$ using homogeneous coordinates. 

\subsection{Motion initialization}
\label{sec:motion_init}
T2V diffusion models provide strong semantic priors for motion synthesis but struggle to directly supervise 3D articulated motion from an uninitialized state. When skeletal parameters are optimized from scratch using SDS, the optimization landscape becomes non-convex and under-constrained, often resulting in degenerate local minima and severe visual artifacts such as floating geometry, foot-skating, jittery motion, and anatomically implausible joint configurations~\cite{bahmani2024tc4d}. To address this limitation, we introduce a motion initialization module that provides a structured, physically plausible prior, significantly stabilizing diffusion-guided optimization.

Given the extracted skeletal hierarchy, we first infer a coarse morphological class of the character: bipedal, quadrupedal, non-living, or flying/aquatic. This is achieved by analyzing the structural properties of the skeletal tree, such as tree depth, branching factor, limb symmetry, and distributions of relative bone lengths. 
In parallel, we parse the input text prompt $y$ using a lightweight semantic classifier to extract the target motion category (e.g., \textit{walking}, \textit{running}, or \textit{jumping}). The combination of morphological type and semantic action forms a structured query that is mapped to procedural kinematic rules. This mapping ensures that the initialized motion respects both the character's physical structure and the intended behavior, thereby assigning a range of motion to each skeletal joint.

To initialize the skeletal kinematics, we generate a dense, frame-by-frame trajectory for the joint rotations $\tilde{\mathbf{R}} \in \mathbb{R}^{T \times J \times 3}$ and root translation $\tilde{\mathbf{t}}_{\text{root}} \in \mathbb{R}^{T \times 3}$ using parameterized periodic functions, where $T$ is the total number of frames and $J$ is the total number of joints. Drawing inspiration from classic procedural animation and Fourier-based kinematic modeling~\cite{unuma1995fourier}, we model the joint rotations for cyclic motions (such as walking or running) using sinusoidal primitives:
\begin{equation}
    \tilde{\mathbf{R}}_{t,j} = \mathbf{A}_j \sin(2\pi \omega t + \boldsymbol{\varphi}_j) + \mathbf{b}_j,
\end{equation}
where $\omega$ is the global temporal frequency dictating the overall cycle speed, while $\mathbf{A}_j$ (amplitude), $\boldsymbol{\varphi}_j$ (phase offset), and $\mathbf{b}_j$ (resting bias) are joint-specific parameters. 
These joint-specific parameters are explicitly retrieved from a predefined heuristic dictionary, which is mapped to the semantic action extracted from the user's prompt. This procedural generation naturally encodes key biomechanical behaviors. For example, assigning a phase difference of $\pi$ between the left and right hip joints enforces phase-shifted limb coordination, while specific amplitude and bias tunings ensure periodic foot-ground contact and pelvis-spine counter-rotation. For non-cyclic or transient motions (e.g., jumping or striking), we utilize piecewise parametric curves with phase-dependent parameter scheduling to model standard acceleration and deceleration patterns.

While the dense trajectories $\tilde{\mathbf{R}}$ provide a meaningful initialization, directly optimizing in the high-dimensional space $\mathbb{R}^{T \times J \times 3}$ leads to temporal instability under SDS gradients. To implement temporal smoothness and reduce optimization complexity, we project these dense trajectories onto a low-dimensional manifold using Non-Uniform Rational B-Splines (NURBS)~\cite{farin1990curves}.
For each joint $j$, we uniformly sample the dense trajectory $\tilde{\mathbf{R}}_{:,j}$ to extract $K+1$ control points $\mathbf{c}_{j,k} \in \mathbb{R}^3$. We then represent the continuous rotational trajectory $\mathbf{R}_j(s)$ over normalized time $s \in [0,1]$ as:
\begin{equation}
    \mathbf{R}_j(s) = \frac{\sum_{k=0}^{K} \bar{B}_{k,d}(s) w_{j,k} \mathbf{c}_{j,k}}{\sum_{k=0}^{K} \bar{B}_{k,d}(s) w_{j,k}},
\end{equation}
where $\bar{B}_{k,d}(s)$ denotes the standard B-spline basis function of degree $d$ (we use cubic splines, $d=3$), and $w_{j,k} \in \mathbb{R}^+$ are the strictly positive rational weights assigned to each control point. It provides a compact, differentiable representation that inherently enforces temporal smoothness and reduces high-frequency artifacts during optimization.

A key advantage of the NURBS formulation is the inclusion of the rational weights $w_{j,k}$, which are critical for modeling non-uniform kinematic transitions. We assign specific weights based on the anatomical function of the joint and the phase of the motion. Specifically, during foot-contact phases at lower-extremity joints, we assign significantly larger weights ($w_{j,k} \gg 1$) to the corresponding control points. 
This effectively pulls the interpolated curve closer to these specific points, creating the sharp transitions necessary to capture the impact dynamics of foot-ground interaction.
In contrast, during swing phases, we assign uniform weights ($w_{j,k} = 1$) to control points to encourage smooth, pendular motion. Similarly, torso and spine joints are assigned moderate weights to preserve stability while allowing subtle compensatory motion. The final continuous trajectories are evaluated via the standard Cox-de Boor recursion algorithm~\cite{cox1972basis} to yield the initialized discrete global root translations $\mathbf{t}_{\text{root}}^{(0)}$ and joint rotations $\mathbf{R}^{(0)}$. The localized translation offsets (discussed in Sec.~\ref{sec:fk}) are initialized strictly to zero ($\boldsymbol{\delta}^{(0)} = \mathbf{0}$). By conditioning the optimization loop on these low-dimensional NURBS parameters, our method ensures that the T2V diffusion model refines a physically plausible, structurally aligned motion prior rather than hallucinating kinematics from random noise.  This significantly improves convergence stability, reduces artifacts, and leads to more physically consistent and semantically aligned animations.

\subsection{Forward Kinematics and skinning}
\label{sec:fk}
To render the final animation, the learned local motion parameters, root trajectory, joint rotations, and non-rigid offsets should be mathematically mapped into global world coordinates. We use a differentiable forward kinematics layer to propagate local motion parameters into global transformations. First, axis-angle rotations $\mathbf{R}_{t,j}$ are converted into rotation matrices $\mathbf{M}_{t,j} \in SO(3)$. We then construct a local transformation matrix $\mathbf{L}_{t,j} \in \mathbb{R}^{4 \times 4}$ for each joint. We incorporate the learnable local offsets $\boldsymbol{\delta}_{t,j}$ directly into the translation component of this transformation to model secondary motion:
\begin{equation}
\mathbf{L}_{t,j} = \begin{bmatrix} \mathbf{M}_{t,j} & \mathbf{o}_{j} + \boldsymbol{\delta}_{t,j} \\
\mathbf{0} & 1 
\end{bmatrix}.
\end{equation}
The global transformation $\mathbf{G}_{t,j}$ for each joint is computed recursively by traversing the pre-computed topological order:
\begin{equation}
\mathbf{G}_{t,j} =
\begin{cases}
\mathcal{T}(\mathbf{t}_{\text{root}}) \cdot \mathbf{L}_{t,j}, 
& \text{if } j \text{ is the root}, \\
\mathbf{G}_{t,p(j)} \cdot \mathbf{L}_{t,j}, 
& \text{otherwise}.
\end{cases},
\end{equation}
where $\mathcal{T}(\mathbf{t}_{\text{root}})$ applies the global root translation. This formulation ensures structural consistency while remaining fully differentiable with respect to $\mathbf{R}$, $\mathbf{t}_{\text{root}}$, and $\boldsymbol{\delta}$.

We use Linear Blend Skinning (LBS) to deform the surface mesh according to the computed skeletal motion. The mesh consists of $V$ vertices with rest positions $\mathbf{v}_i \in \mathbb{R}^3$. Each vertex is associated with a set of skinning weights $w_{i,j}$ satisfying $\sum_j w_{i,j} = 1$, which determine the influence of each joint $j$ on vertex $i$. These weights are pre-computed using puppeteer~\cite{song2025puppeteer} to ensure smooth deformations. 
Let $\mathbf{B}_j = (\mathbf{G}_{j}^{\text{rest}})^{-1}$ denote the inverse bind matrix, which transforms points from the global rest pose into the local coordinate frame of joint $j$. The deformed position $\mathbf{v}_{t, i}$ of vertex $i$ at time $t$ is computed as a weighted sum of the transformed rest positions:
\begin{equation}
\mathbf{v}_{t, i} = \sum_{j=1}^{J} w_{i,j} \left( \mathbf{G}_{t,j} \cdot \mathbf{B}_j \cdot \mathbf{v}_i \right) .
\end{equation} 
By incorporating the local offsets $\boldsymbol{\delta}$ into $\mathbf{G}_{t,j}$ before skinning, our formulation enables the optimization process to learn subtle, non-rigid surface deformations that blend seamlessly across the mesh.

\subsection{Diffusion-Based Motion-aware Score Distillation Guidance}
\label{sec:sds}
To incorporate high-level semantic motion information, we use a pre-trained T2V diffusion model~\cite{wang2023modelscope} as a motion prior. SDS operates by rendering a sequence of images and independently adding gaussian noise to each frame. The diffusion model then predicts the noise per frame. Consequently, the appearance gradients are temporally uncorrelated; the diffusion model may hallucinate disparate textures, colors, or geometric topologies across adjacent frames to satisfy the text prompt. When these independent gradients are backpropagated to the skeletal and local offset parameters, it results in high-frequency surface collapses and structural flickering.
Given a sequence of rendered frames $\mathbf{r} \in \mathbb{R}^{T \times C \times H \times W}$ and a text prompt $y$, we guide the animation using video motion-aware Score Distillation (MoSDS), inspired by Animus3D~\cite{sun2025animus3d}. The diffusion model components, including the text encoder and UNet, remain frozen throughout the optimization process. This ensures that the rich semantic knowledge encapsulated in the pre-trained weights is preserved while providing gradients that encourage the rendered animation to align with the text-conditioned motion distribution. 
To maintain stability and focus on motion dynamics rather than high-frequency noise or structural hallucination, we restrict the diffusion timestep sampling range to $\tau \in [0.02, 0.50]$ of the total training steps. The MoSDS process operates in the latent space of the diffusion model. Rendered frames $\mathbf{r}$ are first resized to $256 \times 256$, normalized to the range $[-1, 1]$, and encoded into a latent video sequence $\mathbf{z} = \mathrm{E}(\mathbf{r})$. Gaussian noise $\epsilon \sim \mathcal{N}(\mathbf{0}, \mathbf{I})$ is added at a randomly sampled timestep $\tau$, producing noisy latents $\mathbf{z}_\tau$. The UNet then predicts the noise residual $\hat{\epsilon}_\theta(\mathbf{z}_\tau; \tau, y)$ conditioned on the text prompt. We use Classifier-Free Guidance (CFG) with a scale of $w_{\text{cfg}}=10.0$ to strengthen the fidelity to the text prompt, computing the final noise prediction as $\hat{\epsilon}_{\text{total}} = \hat{\epsilon}_{\text{uncond}} + \bar{w} (\hat{\epsilon}_{\text{text}} - \hat{\epsilon}_{\text{uncond}})$. 

Unlike SDS, which applies independent gradient updates per frame and frequently destroys temporal coherence, MoSDS decouples the raw noise residual $\Delta \epsilon = \hat{\epsilon}_{\text{total}} - \epsilon$ along the temporal axis. We extract the appearance gradient $\Delta\epsilon_{\text{appear}}$ by computing the temporal mean of the noise across all $N$ frames, and the motion gradient $\Delta\epsilon_{\text{motion}}$ as the frame-wise deviation from this mean, such that $\Delta\epsilon_{\text{appear}} = \frac{1}{N} \sum_{t=1}^{N} \Delta \epsilon_t$ and $\Delta\epsilon_{\text{motion}} = \Delta \epsilon -\Delta\epsilon_{\text{appear}}$.
The final MoSDS gradient with respect to the latent is formulated as a weighted combination of these decoupled signals:
\begin{equation}
\nabla_{\mathbf{z}} \mathcal{L}_{\text{MoSDS}} = \bar{w}(\tau) \left( \lambda_{\text{appear}} \Delta\epsilon_{\text{appear}} + \lambda_{\text{motion}} \Delta\epsilon_{\text{motion}} \right),
\end{equation}
where $\bar{w}(\tau)$ depends on the noise schedule, and $\lambda_{\text{appear}}, \lambda_{\text{motion}}$ are scalar weights. By heavily weighting $\lambda_{\text{motion}}$ and suppressing $\lambda_{\text{appear}}$, we force the optimizer to utilize the diffusion model strictly as a kinematic critic, updating the underlying skeletal parameters based on motion variance while preserving the 3D geometry.

To propagate these gradients back to the skeletal parameters, we use a proxy loss directly in the latent space~\cite{poole2022dreamfusion}. We treat the MoSDS gradient as a constant and construct a target latent sequence $\mathbf{z}_{target} = \mathbf{z} - \eta \nabla_{\mathbf{z}} \mathcal{L}_{\text{MoSDS}}$. A Mean Squared Error (MSE) loss is computed between the encoded latents $\mathbf{z}$ and $\mathbf{z}_{target}$. Minimizing this proxy loss allows smooth gradients to flow backward through the frozen VAE encoder and the differentiable renderer, directly updating the skeletal kinematics and local offsets.

\subsection{Physics-Inspired Motion Regularization}
\label{sec:physics}
While diffusion guidance provides strong semantic priors, it struggles to ensure temporally stable and physically plausible motion. 
To ensure the generated animation follows physics-inspired constraints and exhibits temporal coherence, we minimize a comprehensive set of physics-based regularization terms. This objective is crucial for preventing artifacts such as jitter, foot sliding, and unnatural joint rotations.

\paragraph*{Temporal Smoothness}
We implement motion continuity by using hierarchical derivatives of the kinematic parameters to simulate inertia and damping~\cite{witkin1988spacetime, holden2016deep}.  The velocity loss ($\mathcal{L}_{vel}$) penalizes the first-order temporal differences of both joint rotations and root translations. By minimizing the squared $L_{2}$ norm of the velocity, we prevent high-frequency jitter and rapid, unnatural pose changes between consecutive frames. Furthermore, we add an acceleration loss ($\mathcal{L}_{accel}$) that minimizes the second-order differences of the motion parameters. This term ensures smooth velocity transitions, effectively adding simulated weight to the character and preventing the weightless float often seen in generated animations. Let $\Phi = \{\Phi_t\}_{t=1}^{T}$ denote the sequence of kinematic parameters, including joint rotations and root translations.
The loss is defined as the mean squared norm of the velocity $\Delta \Phi_t = \Phi_t - \Phi_{t-1}$. The unified hierarchical smoothness loss is defined as:
\begin{equation}
\begin{aligned}
\mathcal{L}_{\text{smooth}} = \;&
\lambda_{\text{vel}} \left( \frac{1}{T-1} \sum_{t=2}^{T} \|\Delta {\Phi}_t\|_2^2 \right) \\
&+ \lambda_{\text{accel}} \left( \frac{1}{T-2} \sum_{t=3}^{T} \|\Delta^2 {\Phi}_t\|_2^2 \right),
\end{aligned}
\end{equation}
where $\lambda_{\text{vel}}$ and $\lambda_{\text{accel}}$ control the relative strength of velocity and  acceleration respectively.

\paragraph*{Topological Constraints}
To prevent the optimization from producing unnatural poses, we apply soft constraints based on the character's skeletal structure, inspired by SMPLify~\cite{bogo2016keep}. We define joint limits ($\mathcal{L}_{\text{rom}}$) by classifying joints into anatomical categories and assigning specific rotational thresholds $\theta_{\text{limit}}$. For instance, joints identified as part of the spine are assigned stricter rotational limits (e.g., 0.4 radians) to maintain torso stability. In contrast, hinge-like joints such as knees or elbows are permitted a wider range of motion (e.g., 1.5 radians). We utilize a ROM loss function that penalizes the magnitude of rotations $\mathbf{R}_{t,j}$ only when they exceed these predefined limits, such as:
\begin{equation}
  \mathcal{L}_{\text{rom}} = \frac{1}{TJ} \sum_{t=1}^{T} \sum_{j=1}^{J} \max(0, (\|\mathbf{R}_{t,j}\|_2 - \theta_{\text{limit}, j}))^2.
\end{equation}
ROM loss function allows free motion within the natural range while heavily penalizing hyperextension.

Additionally, we enforce symmetry loss ($\mathcal{L}_{\text{sym}}$) for characters with bilateral structures. By identifying paired joints (e.g., left and right legs), we minimize the difference in motion magnitude between corresponding joints $\mathrm{\bar{C}}$. This regularization prevents asymmetric gaits where one limb moves significantly more than the other, ensuring balanced and coordinated locomotion~\cite{holden2016deep} such as:
\begin{equation}
    \mathcal{L}_{\text{sym}} = \frac{1}{T |\mathrm{\bar{C}}|} \sum_{t=1}^{T} \sum_{(\text{left}, \text{right}) \in \mathrm{\bar{C}}} (\|\mathbf{R}_{t, \text{left}}\|_2 - \|\mathbf{R}_{t,\text{right}}\|_2)^2.
\end{equation}

\paragraph*{Cyclic Consistency}
For animations intended to loop seamlessly, such as walk or run cycles, we enforce cyclic consistency ($\mathcal{L}_{\text{cyclic}}$) via a boundary constraint. This loss minimizes the $L_{2}$ distance between the pose (joint rotations and root translations) of the first frame ($t=1$) and the last frame ($t=T$), inspired by~\cite{starke2022deepphase}. Beyond positional matching, we also constrain the velocity at the boundaries to match, ensuring a seamless transition from the end of the loop back to the start: 
\begin{equation}
    \mathcal{L}_{\text{cyclic}} = \| \Phi_1 - \Phi_T \|_2^2 + \| \Delta \Phi_1 - \Delta \Phi_T \|_2^2.
\end{equation}

The combined physics-based regularization is formulated as:
\begin{equation}
\begin{aligned}
\mathcal{L}_{phy} =\;&
\lambda_{\text{smooth}} \mathcal{L}_{\text{smooth}} +
\lambda_{\text{rom}} \mathcal{L}_{\text{rom}} \\
&+
\lambda_{\text{cyclic}} \mathcal{L}_{\text{cyclic}} + 
\lambda_{\text{sym}} \mathcal{L}_{\text{sym}}, \\
\end{aligned}
\end{equation}
where, $\lambda_{\text{smooth}}$, $\lambda_{\text{rom}}$, $\lambda_{\text{cyclic}}$ and $\lambda_{\text{sym}}$ denotes the corresponding weighting coefficients.

\begin{algorithm}[h!]
\LinesNumbered
\caption{Diffusion-Guided Skeleton-based Text-to-3D Animation}
\label{alg:sketchanim_updated}
\SetAlgoLined
\KwIn{
    2D sketch $S$ and text prompt $y$
}
\KwOut{
    Optimized joint rotations $\{\mathbf{R}_{t,j}\}$, root translations $\{\mathbf{t}_{\text{root}, t}\}$, local offsets $\{\boldsymbol{\delta}_{t,j}\}$, and final deformed mesh sequence $\mathbf{V}_{\text{seq}}$
}
\textbf{Step 1: 3D Initialization}\\
Estimate 3D mesh $\mathbf{V}^0$ from 2D sketch $S$, extract skeleton hierarchy $\mathcal{H}$ and skinning weights $\mathbf{w}$. 

\textbf{Step 2: Motion Initialization} \\
$\{\mathbf{R}_{t,j}, \mathbf{t}_{\text{root}, t}, \boldsymbol{\delta}_{t,j}\} \leftarrow \text{MotionInitializer}(\mathcal{H}, y, T)$ \tcp*{Initialize base motion (NURBS)}
$\Psi \leftarrow \text{InitSpringMassSim}(\mathbf{V}^0, \mathcal{D}_{\text{mask}})$ \tcp*{$\mathcal{D}_{\text{mask}}\rightarrow$ Dynamic Region Mask}

\ForEach{iteration $u=1,\dots,M$}{
    \textbf{Step 3: Forward Kinematics with Local Offsets} \\
    \ForEach{frame $t=1,\dots,T$}{
        \ForEach{joint $j \in \mathcal{H}$ (topological order)}{
            $\mathbf{G}_{t,j} \leftarrow \mathbf{G}_{t,\text{p}(j)} \cdot \mathcal{T}(\mathbf{R}_{t,j}, \mathbf{o}_j + \boldsymbol{\delta}_{t,j})$
        }
    }

    \textbf{Step 4: Mesh Skinning \& Secondary Dynamics} \\
    \ForEach{frame $t=1,\dots,T$}{
        \ForEach{joint $j \in \mathcal{H}$}{
            $\mathbf{M}_{t,j} \leftarrow \mathbf{G}_{t,j} \mathbf{B}_j^{-1}$
        }
        $\mathbf{V}^{\text{LBS}}_t \leftarrow \sum_{j} \mathbf{w}_j \mathbf{M}_{t,j} \mathbf{V}^0$ \tcp*{LBS}
    }
    $\mathbf{V}_{\text{seq}} \leftarrow \Psi(\mathbf{V}^{\text{LBS}}, \mathcal{D}_{\text{mask}})$ \tcp*{Spring-Mass force for secondary motion effects}

    \textbf{Step 5: Differentiable Rendering \& MoSDS Guidance} \\
    Render $\mathbf{V}_{\text{seq}}$ to generate 2D frames $\textbf{r}$. \\
    Compute MoSDS gradient via decoupled appearance and motion noise residuals:
    \[
    \nabla_{\text{MoSDS}} \leftarrow \lambda_{\text{appear}} \Delta\epsilon_{\text{appear}} + \lambda_{\text{motion}} \Delta\epsilon_{\text{motion}} 
    \]
    \textbf{Step 6: Physics Regularization} \\
    Physics-based losses ($\mathcal{L}_{\text{phy}}$), and physical contact ($\mathcal{L}_{\text{env}}$).

    \textbf{Step 7: Parameter Update} \\
    Minimize total progressive loss utilizing iteration-dependent weights $\lambda(u)$:
    \[
    \mathcal{L}_{\text{total}} = \lambda_{\text{MoSDS}}(u) \mathcal{L}_{\text{proxy}} + \sum_k \lambda_k(u) \mathcal{L}_k
    \]
    Update $\{\mathbf{R}_{t,j}, \mathbf{t}_{\text{root}, t}, \boldsymbol{\delta}_{t,j}\}$ via AdamW optimizer.
}
\Return{\textbf{Final animated mesh sequence $\mathbf{V}_{\text{seq}}$}}
\end{algorithm}

\paragraph*{Environment interaction}
Grounding the character in a physical environment is essential for realistic animation. We address this through a ground penetration ($\mathcal{L}_{\text{ground}}$) term that strictly penalizes the vertical position $v^{z'}_{t,i}$ of any vertex $i$ that falls below the ground plane ($z'=0$)~\cite{tevet2022human, yuan2023physdiff}. This ensures the character remains above the surface and does not clip through the floor: 
\begin{equation}
    \mathcal{L}_{\text{ground}} = \frac{1}{T V} \sum_{t=1}^{T} \sum_{i=1}^{V} \max(0, -v^{z'}_{t,i})^2. 
\end{equation}
Furthermore, to eliminate the common artifact of foot sliding (where feet glide across the ground while holding weight), we use a foot contact ($\mathcal{L}_{\text{contact}}$) loss. This term dynamically detects contact phases by monitoring the height of foot joints; when a foot is near the ground threshold, the loss minimizes its horizontal velocity~\cite{bogo2016keep}. It effectively plants the foot in place during the stance phase of locomotion, simulating proper friction and weight transfer. Let $\mathbf{p}_{f,t} \in \mathbb{R}^3$ denote the 3D position of a designated foot joint $f$ at frame $t$, and define its horizontal velocity
$\mathbf{vel}^{xy}_{f,t} = \left( \mathbf{p}_{f,t} - \mathbf{p}_{f,t-1} \right)_{xy}$. We define a binary contact indicator $h_{f, t} \in \{0,1\}$, which is activated when the foot joint is sufficiently close to the ground as 
$h_{f,t} = \mathbb{I}\big( p^{z}_{f,t} < \tau \big)$, where $p^{z}_{f,t}$ is the vertical coordinate and $\tau$ is a predefined contact threshold.
The contact loss then penalizes horizontal motion during these contact phases:
\begin{equation}
\mathcal{L}_{\text{contact}} 
= 
\frac{1}{\sum_{t=1}^{T} h_{f,t} + \epsilon} 
\sum_{t=1}^{T} 
h_{f,t} \, \| \mathbf{vel}^{xy}_{f,t} \|_2^2,
\end{equation}
where $\epsilon$ is a constant for numerical stability.

All environmental constraints are integrated into the generative optimization process:
\begin{equation}
\begin{aligned}
\mathcal{L}_{\text{env}} = \;&
\lambda_{\text{ground}}\mathcal{L}_{\text{ground}} + \lambda_{\text{contact}} \mathcal{L}_{\text{contact}}  \\
\end{aligned}
\end{equation}

\subsection{Secondary Motion Effects}
To model subtle, non-rigid secondary motions, we integrate learnable per-joint local translations. However, without proper constraints, these offsets can distort the underlying skeletal structure or introduce unrealistic deformations. To address this, we apply offset regularization ($\mathcal{L}_{\text{offset}}$), which penalizes the squared $L_{2}$ norm of the local offsets, constraining them to represent small, detailed deformations rather than large-scale structural changes. Additionally, we penalize the temporal velocity of these offsets to ensure that the secondary motion is temporally coherent and does not flicker or jitter independently of the primary skeletal animation. 
Let $\boldsymbol{\delta}_{t} \in \mathbb{R}^{J \times 3}$ denote the set of local offsets for all joints at frame $t$. The offset regularization loss is defined as:
\begin{equation}
\mathcal{L}_{\text{offset}} = 
\frac{1}{T} \sum_{t=1}^{T} \| \boldsymbol{\delta}_{t} \|_2^2 + \lambda_{\delta} \left( \frac{1}{T-1} \sum_{t=2}^{T} \| \boldsymbol{\delta}_{t} - \boldsymbol{\delta}_{t-1} \|_2^2 \right),
\end{equation}
where $\lambda_{\delta}$ controls the strength of the temporal smoothness regularization.

Before physics-based simulation, the system automatically partitions the mesh into rigid and deformable regions, determining which vertices are driven solely by skeletal motion (rigid) and which are controlled by physics (soft). This classification is derived from the skinning weights $\mathbf{w}$ and skeletal topology, without requiring manual annotations. Vertices with high weight concentration on core structural bones (e.g., spine, skull) are designated as anchor vertices and remain rigidly controlled via LBS, serving as attachment points for the physics system. In contrast, vertices associated with extremities or non-structural joints (e.g., hair, cloth, tail) are classified as dynamic vertices and are simulated as soft bodies. Additionally, the mesh is partitioned into semantically meaningful regions, enabling region-specific masks (e.g., for hair or cloth), which allow the physics module to apply distinct dynamic behaviors to different parts of the character.
Once the vertices are classified, the Spring-Mass Simulator module assigns secondary motion to the dynamic region inspired by~\cite{sifakis2007hybrid, selle2008mass, willett2017secondary}.

For each dynamic vertex $i$ at a given time step, the system computes the total force $\mathbf{F}_i$ acting on the vertex as the sum of a position spring force, a structural spring force, velocity damping, and gravity. Specifically, the position spring force $\mathbf{F}_{\text{spring}} = -k_{\text{pos}}(\mathbf{q}_i - \hat{\mathbf{q}}_i)$ binds the simulated position $\mathbf{q}_i$ to its underlying kinematic LBS target $\hat{\mathbf{q}}_i$, ensuring the soft tissue follows the primary skeletal animation. To prevent volume collapse, the structural spring force $\mathbf{F}_{\text{struct}}$ penalizes deviations from the rest-pose edge lengths $l^0_{ij}$ for adjacent vertices $j$, formulated as:
\begin{align*}
    F_{\text{struct}} = \sum_{j} k_{\text{struct}} \cdot \Delta l_{ij} \frac{q_j - q_i}{\|q_j - q_i\|}, 
\end{align*}
where the stretch $\Delta l_{ij}$ is clamped to a maximum of $30\%$ of the rest length to maintain stability. Finally, a damping force $F_{damp} = -d \cdot vel_i$,
and an gravitational force $F_{grav} = [0, -g \cdot m_i, 0]^\top$, are applied, yielding the total acceleration $a_i = (F_{\text{spring}} + F_{\text{struct}} + F_{\text{damp}} + F_{\text{grav}}) / m_i$. The system state is updated over discrete substeps $\Delta t$ using a semi-implicit Euler integration scheme. Because physics simulations within gradient-based optimization loops are highly susceptible to numerical instability, we enforce strict clamping mechanisms during integration. First, the updated velocity $vel_i \leftarrow vel_i + a_i \Delta t$ is strictly bounded by a maximum magnitude $vel_{max}$ to prevent drastic acceleration spikes, expressed as:
\begin{align*}
    vel_i \leftarrow vel_i \cdot \min(1, \frac{vel_{max}}{\|v_i\|}). 
\end{align*}
Subsequently, the newly integrated position $q_i \leftarrow q_i + vel_i \Delta t$ undergoes displacement clamping, ensuring its Euclidean distance from the LBS target $\hat{q}_i$ never exceeds a predefined threshold $d_{max}$, which strictly prevents mesh tearing. To ensure a seamless visual transition between the simulated extremities and the rigid kinematic core, the final output position $q^{out}_i$ is computed via linear interpolation using a spatially varying blend weight $w_i \in [0, 1]$. This is formulated as $q^{out}_i = \hat{q}_i + w_i (q_i-\hat{q_i})$, allowing the generative MoSDS loss to evaluate the visual quality of the secondary dynamics and backpropagate gradients directly through the simulation to optimize the underlying skeletal parameters.

The final optimization objective is:
\begin{equation}
\begin{aligned}
\mathcal{L}_{total} =\;&
\lambda_{\text{MoSDS}} \mathcal{L}_{\text{MoSDS}} +
\lambda_{\text{phy}} \mathcal{L}_{\text{phy}} \\
&+
\lambda_{\text{env}} \mathcal{L}_{\text{env}} +
\lambda_{\text{offset}} \mathcal{L}_{\text{offset}} \\
\end{aligned}
\end{equation}


\section{Experiment and results}

\begin{figure*}[!htbp]
  \centering
    \includegraphics[width=0.75\linewidth]{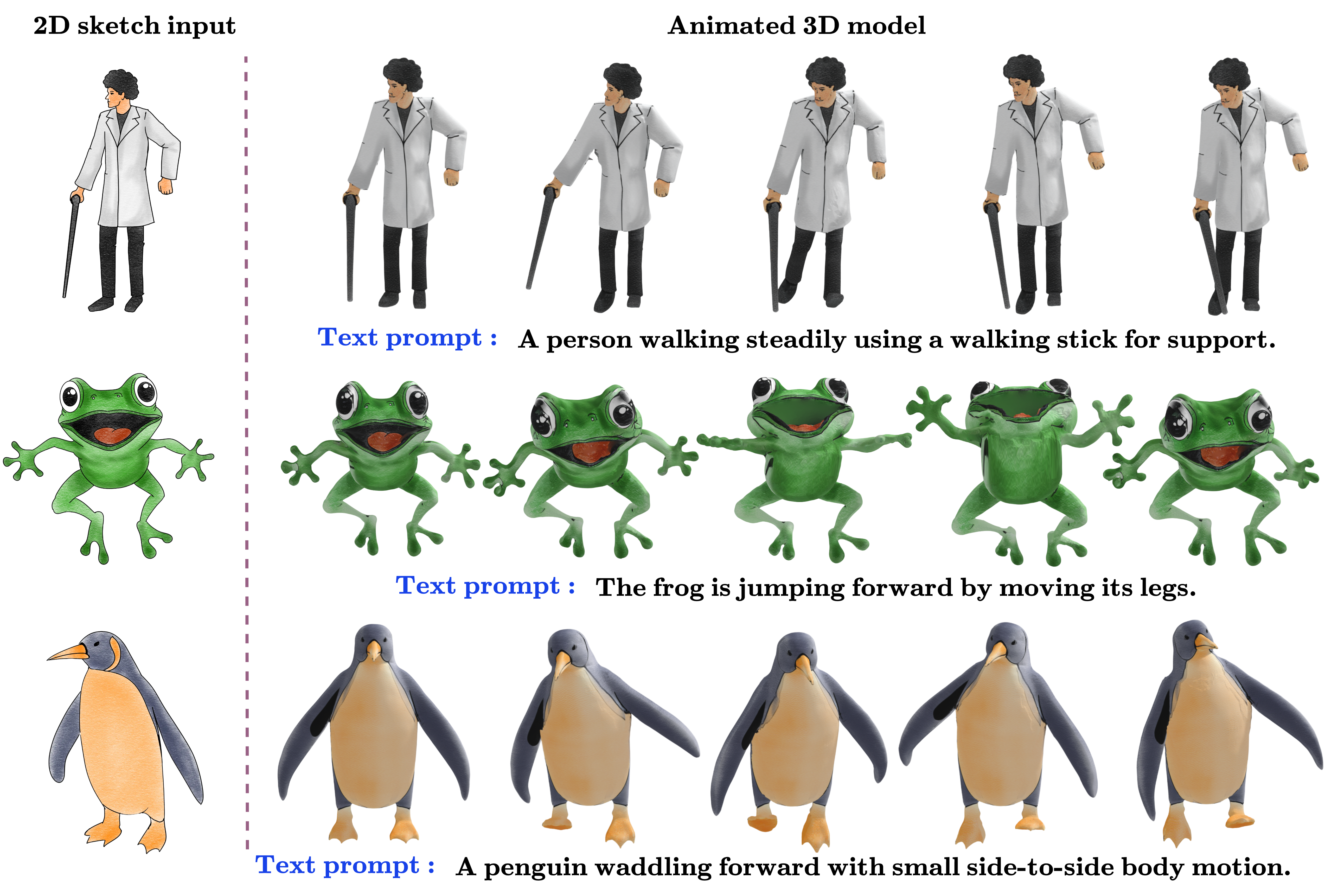}
    \caption{Qualitative results of our proposed method and generated 3D animation sequences from input text prompts.}
    \label{fig:results}
\end{figure*}

\subsection{Implementation details}
Our method uses PyTorch3D for differentiable rendering and geometric operations to optimize text-driven motion for a rigged 3D mesh. We parameterize animation as a sequence of $T=48$ frames with learnable joint rotations ${R}\in\mathbb{R}^{T\times J\times3}$ (axis-angle), root translations ${t}_{\text{root}}\in\mathbb{R}^{T\times3}$, and per-joint local offsets ${\delta}\in\mathbb{R}^{T\times J\times3}$ to model secondary non-rigid motion. Global joint transforms are computed via forward kinematics, and vertices are deformed using linear blend skinning. Text-to-video supervision is provided by the frozen text-to-video diffusion model~\cite{wang2023modelscope} using MoSDS, and CFG scale set to $10.0$; SDS~\cite{poole2022dreamfusion} gradients are computed in latent space and backpropagated via an MSE loss on decoded rendered frames. Optimization uses AdamW~\cite{loshchilov2017decoupled} with weight decay $10^{-5}$, learning rates of $1.5\times10^{-2}$, $2.0\times10^{-2}$, and $5.0\times10^{-3}$ for rotations, root translations, and offsets respectively. It takes around $5$ minutes to generate the final animated sequence.

\subsection{Dataset}
We create a dataset of 100 hand-drawn sketches to evaluate the proposed method's performance. The dataset is designed to cover a diverse range of articulation structures and object types, and is divided into three categories: biped (40 sketches), quadruped (40 sketches), and non-living objects (20 sketches). This categorization allows us to assess the generalization capability of our approach across different kinematic topologies and motion complexities, ranging from articulated characters to rigid or semi-rigid objects. For each sketch, we use Step1X-3D~\cite{li2025step1x} to inflate the 2D drawing into a textured 3D mesh representation. The resulting meshes serve as canonical 3D representations for subsequent animation. We then apply Puppeteer~\cite{song2025puppeteer} to extract skeletal structures and corresponding skinning weights from the reconstructed meshes automatically. The extracted skeletons, together with the textured meshes, are subsequently used as inputs to our diffusion-guided animation pipeline. This dataset and evaluation setup enable a comprehensive analysis of motion quality, structural consistency, and semantic alignment across diverse sketch categories.

\begin{table}[ht]
\centering
\begin{threeparttable}[b]
\caption{Quantitative comparison of our approach with state-of-the-art methods.}
\label{tab:comp_table}
\setlength{\tabcolsep}{4pt}
\small
\begin{tabular}{lccc}
\toprule
\textbf{Methods} &
\textbf{T2VA} ($\uparrow$) &
\textbf{MLD} ($\downarrow$) &
\textbf{MVFC} ($\uparrow$) \\
\midrule
AnimateAnymesh~\cite{wu2025animateanymesh} & 0.1967 & 0.6796 & 0.8810 \\
BiMotion~\cite{wang2026bimotion} & 0.1891 & 0.0102\tnote{*} & 0.8902 \\
\textbf{Ours} & \colorbox{colorFst}{0.2120} & \colorbox{colorFst}{0.3523} & \colorbox{colorFst}{0.8933} \\
\bottomrule
\end{tabular}
\begin{tablenotes}
    \item [*] \tiny Such a low value is a result of negligible motion in the mesh sequence.
\end{tablenotes}
\end{threeparttable}
\end{table}

\subsection{Evaluation matrices}
We use the following metrics for quantitative evaluation of the proposed method's performance.

\paragraph*{Text-to-video alignment (T2VA) ($\uparrow$)}
Text-to-video alignment evaluates the semantic consistency between the input text prompt and the generated animation sequence.  This is achieved using X-CLIP~\cite{ma2022x}, text embedding $\phi_{\text{text}}$, and frame embeddings $\Theta_{\text{frame}}$ are projected into a shared embedding space, and their similarity is measured using cosine similarity:
\[
\hat{S}_{\text{X-CLIP}} = \frac{1}{T} \sum_{t=1}^{T} \frac{\langle\phi_{\text{text}}, \Theta_{\text{frame}}\rangle}{\|\phi_{\text{text}}\| \|\Theta_{\text{frame}}\|},
\]
where $\hat{S}_{\text{X-CLIP}}$ denotes the cosine similarity and $\langle\cdot,\cdot\rangle$ denotes the inner product. A higher score indicates better semantic alignment between the prompt and the generated video frames.

\paragraph*{Mesh Laplacian Distortion (MLD) ($\downarrow$)}
Mesh Laplacian Distortion measures the geometric stability of a deforming mesh by preserving the local structural relationship between vertices and their neighbors. For a vertex $v_i$ with neighbors $n(i)$, the Laplacian coordinate is defined as $\rho_i = v_i - \frac{1}{|n(i)|}\sum_{\tilde{k} \in n(i)} v_{\tilde{k}}$. The distortion between the deformed mesh and the rest pose is computed as:
\[
E_{\text{lap}} = \sum_{i=1}^{V_{\text{total}}} \left\| \rho_i - \bar{\rho}_i \right\|_2^2,
\]
where lower values indicate better preservation of surface structure during deformation.

\paragraph*{Multi-View Feature Consistency (MVFC) ($\uparrow$)}
Multi-view feature consistency evaluates the 3D coherence of the generated animation by comparing features extracted from renderings of the mesh across different viewpoints. 
Let $\hat{M}_t$ represent the predicted 3D mesh at time frame $t$, and let $\hat{R}_k$ denote the rendering function configured for the $k$-th camera viewpoint, out of $K$ total views. Given $\mathcal{K}$ camera views, features are extracted using a pre-trained encoder $\mu$ (CLIP), and the consistency is measured as the average pairwise cosine similarity:
$$C_{MV}=\frac{1}{\mathcal{K}(\mathcal{K}-1)}\sum_{k\neq k'}\frac{\langle \mu(\hat{R}_k(\hat{M}_t)),\mu(\hat{R}_{k'}(\hat{M}_t))\rangle}{\|\mu(\hat{R}_k(\hat{M}_t))\|\|\mu(\hat{R}_{k'}(\hat{M}_t))\|} ,$$
Higher scores indicate stronger alignment across the rendered views, reflecting a more coherent and geometrically consistent 3D structure.

\subsection{Comparison}
Our method generates suitable and text-aligned motion, as illustrated in Figure~\ref{fig:results}. Furthermore, we compare our approach with state-of-the-art methods for text-driven 3D animation, including AnimateAnyMesh~\cite{wu2025animateanymesh} and BiMotion~\cite{wang2026bimotion}. Although AKD~\cite{li2025articulated} and AnimaX~\cite{huang2025animax} are also relevant skeleton-driven approaches, their official implementations are not publicly available, preventing direct experimental comparison. Therefore, we evaluate our method against the above baselines, which span diverse paradigms in text-driven 3D animation and provide a comprehensive benchmark for assessing our method.

\begin{figure*}
  \centering
    \includegraphics[width=\linewidth]{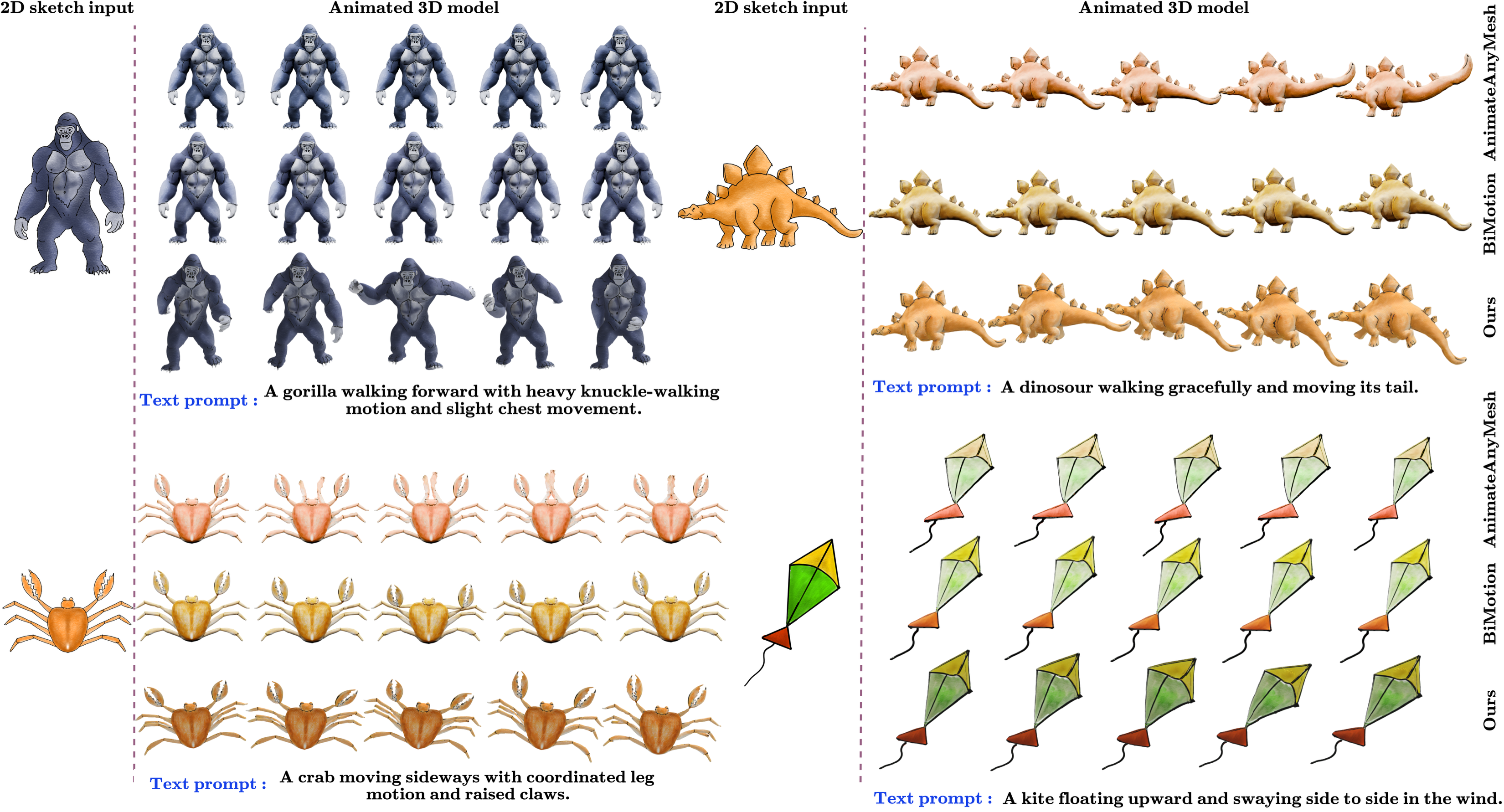}
    \caption{Qualitative Comparison with state-of-the-art methods AnimateAnyMesh~\cite{wu2025animateanymesh}  and BiMotion~\cite{wang2026bimotion}).}
    \label{fig:comparison}
\end{figure*}

\subsubsection{Qualitative Comparison}
We qualitatively compare our approach with state-of-the-art baselines: AnimateAnyMesh~\cite{wu2025animateanymesh}, and BiMotion~\cite{wang2026bimotion}, visual results provided in Figure~\ref{fig:comparison}. 
AnimateAnyMesh~\cite{wu2025animateanymesh} enables motion generation on arbitrary meshes by leveraging diffusion priors and deformation fields. While it significantly improves generalization across diverse mesh topologies, it primarily operates in a deformation space and does not enforce explicit skeletal or kinematic constraints. Consequently, it produces temporally smooth but structurally inconsistent motions shown in Figure~\ref{fig:comparison}, particularly for articulated objects where joint coherence and bone-length preservation are critical. Although this method is effective for subtle movements, its approach does not generalize well to large articulated movements or complex full-body animation. Meanwhile, BiMotion~\cite{wang2026bimotion} introduces a bidirectional motion representation that jointly models motion dynamics and geometry for text-driven animation. Although it improves motion diversity and temporal coherence compared to prior diffusion-based approaches, it still relies on implicit motion representations and lacks strong structural priors for articulated bodies. This leads to ambiguity in joint behavior and reduced controllability of the complex actions shown in Figure~\ref{fig:comparison}. However, these methods are trained on specific datasets; thus, they struggle to generalize to novel articulated meshes and diverse text prompts. In contrast to these baselines, our method operates directly on explicit skeletal parameters and 3D meshes. By extending diffusion guidance to full articulated skeletons and integrating forward kinematics with physics-inspired regularization, our framework ensures physically plausible articulation. This explicit parameterization provides precise control over complex motion semantics, including secondary effects, while maintaining geometric consistency and stable deformation across all frames and viewpoints.

\subsubsection{Quantitative Comparison}
Table~\ref{tab:comp_table} presents a quantitative comparison of our method with AnimateAnyMesh~\cite{wu2025animateanymesh} and BiMotion~\cite{wang2026bimotion} across three complementary metrics: text-to-video alignment, mesh Laplacian distortion, and multi-view feature consistency. These metrics collectively evaluate semantic alignment, geometric stability, temporal smoothness, and 3D coherence. Lower values indicate better performance for mesh distortion, while higher values are preferred for text alignment and multi-view consistency.
AnimateAnyMesh~\cite{wu2025animateanymesh} improves over prior diffusion-based approaches by leveraging deformation priors, resulting in moderate improvements in geometric stability and smoother motion. However, due to the absence of explicit skeletal constraints, it exhibits higher mesh distortion and lower multi-view feature consistency. BiMotion~\cite{wang2026bimotion} struggles to generate dynamic animations, often producing small movements. Due to the limited motion, Laplacian distortion is very low; nevertheless, its implicit motion representation limits global 3D consistency, resulting in comparatively lower text-to-video alignment.
In contrast, our method consistently outperforms both baselines across all metrics. We achieve the lowest mesh Laplacian distortion, indicating stable, smooth surface deformation. Additionally, our approach attains higher text-to-video alignment, reflecting stronger semantic consistency between the generated animation and the input prompt. Finally, the significant improvement in multi-view feature consistency confirms that our framework produces coherent 3D motion representations that remain stable across varying viewpoints.

Overall, quantitative and qualitative results demonstrate that our method achieves better semantic motion alignment, lower mesh distortion, and multi-view feature consistency compared to all baselines. These improvements highlight the advantages of combining diffusion-based semantic guidance with explicit skeletal modeling and physics-aware regularization.

\subsection{Ablation study}
To validate the effectiveness of our technique, we conduct an ablation study by systematically removing key components of our pipeline. Figure~\ref{fig:ablation} presents the qualitative results, while Table~\ref{tab:abl_table} reports the quantitative analysis under different settings. We evaluate the impact of motion initialization, physics-based regularization terms, and secondary motion effects on the generated animations from both qualitative and quantitative perspectives.

\subsubsection{Without Motion Initialization}
\begin{figure}
  \centering
    \includegraphics[width=\linewidth]{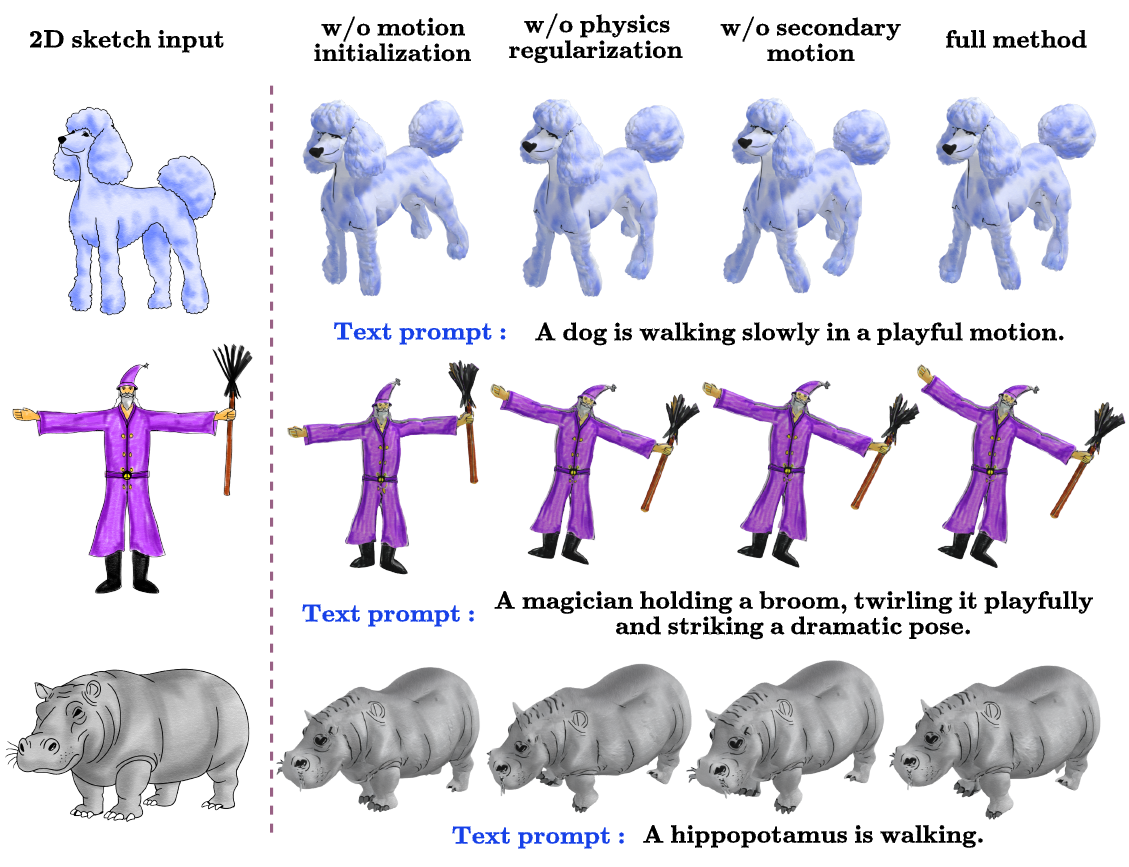}
    \caption{Visual results of ablation study on different settings such as w/o motion initialization, w/o physics regularization, w/o secondary motion, and our full method.}
    \label{fig:ablation}
\end{figure}

To validate the necessity of our procedural motion initialization module (Section~\ref{sec:motion_init}), we evaluate a baseline (\textit{w/o Motion Init}) where all skeletal parameters, joint rotations, root translations, and local offsets are zero-initialized to a static rest pose across all frames. 
In this setting, Video MoSDS synthesizes the full kinematic trajectory from scratch, which often leads to poor local minima due to the lack of inherent 3D structural awareness in text-to-video diffusion models. Consequently, the optimization exhibits artifacts such as floating locomotion, excessive reliance on root translation, foot-skating, and physically implausible joint configurations. Additionally, direct optimization in the high-dimensional space introduces significant temporal jitter, degrading motion smoothness. The limitations of the zero-initialized baseline are clearly visible in Figure~\ref{fig:ablation} and Table~\ref{tab:abl_table}. In contrast, our full pipeline utilizes a low-dimensional NURBS projection of a procedural, biomechanically informed base motion, initializing the parameters within a physically plausible kinematic manifold. As a result, enabling video MoSDS to act as a semantic refiner, adaptively modifying the base gait to align with the text prompt while preserving critical phase dynamics.
This leads to improved joint motion consistency and significantly faster convergence, demonstrating that a strong morphological prior effectively regularizes the highly non-convex MoSDS optimization landscape.

\begin{table}[ht]
\centering
\caption{Quantitative ablation of our approach, evaluating the impact w/o motion initialization, w/o physics-based regularization, and w/o secondary motion, compare to our full method.}
\label{tab:abl_table}
\scalebox{0.9}{
\begin{tabular}{l c c c c}
\toprule
\textbf{Methods \textbackslash Metrics} &
\textbf{T2VA} ($\uparrow$) &
\textbf{MLD} ($\downarrow$) &
\textbf{MVFC} ($\uparrow$) \\
\midrule
w\textbackslash o motion initialization & 0.2059 & 0.4241 & 0.8930 \\
w\textbackslash o physics-based regularization & 0.2071 & 0.3745 & 0.8920 \\
w\textbackslash o secondary motion  & 0.2089 & \colorbox{colorFst}{0.3195} & 0.8927  \\
\textbf{full method} & \colorbox{colorFst}{0.2120} & 0.3523 & \colorbox{colorFst}{0.8933} \\
\bottomrule
\end{tabular}
}
\end{table}

\subsubsection{Without physics-based regularization}
In this setting, we use the skeletal representation without the physics-based regularization terms from the objective function. We set the weights for smoothness, topological constraints, ground contact, and symmetry to zero, relying solely on the MoSDS loss for guidance. Removing physics-based priors results in motion that is visually plausible in static frames but physically incoherent over time. We observe several specific modes of failure: Without the temporal smoothness terms ($\mathcal{L}_{\text{vel}}, \mathcal{L}_{\text{acc}}$), the animation suffers from high-frequency flickering and erratic pose changes, shown in Figure~\ref{fig:ablation}. Without foot contact loss ($\mathcal{L}_{\text{contact}}$) and ground penetration penalty ($\mathcal{L}_{\text{ground}}$), the character exhibits motion artifacts. Feet slide across the base during the stance phase instead of remaining planted, and limbs frequently clip through the ground plane, breaking the illusion of physical interaction. Without joint limit constraints ($\mathcal{L}_{\text{rom}}$), the optimization exploits the full range of rotation to satisfy the visual prompt, resulting in broken joints (e.g., backward bending of the knees) and biomechanically impossible poses. Further, removing the symmetry regularization ($\mathcal{L}_{\text{sym}}$) for bilateral characters leads to uncoordinated locomotion as shown in Figure~\ref{fig:ablation}, with significant differences in limb motion magnitudes, resulting in a limping or lopsided appearance. Table~\ref{tab:abl_table} provides a quantitative analysis of this configuration in terms of text-to-video alignment, mesh Laplacian distortion, and multi-view feature consistency. These results confirm that while SDS provides necessary semantic guidance, physics-based regularization is essential for grounding the generated motion in biomechanical reality.

\subsubsection{Without secondary motion}
In this configuration, we remove the secondary dynamics module, leaving the mesh to rely strictly on the primary skeletal deformation via LBS. Vertices classified as dynamic soft bodies (e.g., hair or cloth) are treated as rigid extensions of the bone hierarchy. While the primary joints successfully execute the text-guided action, the resulting animation appears stiff and artificial, resembling a rigid figure rather than a compliant, organic entity. We observe a distinct lack of "follow-through" and "overlapping action," fundamental principles of natural movement shown in Figure~\ref{fig:ablation}. For instance, a dog's tail and the magician's cap fail to exhibit the natural phase-delayed lag and spring-mass recoil dictated by momentum and gravity. By skipping these secondary spring-mass dynamics, the system loses the subtle, passive physical reactions that are essential for bridging the gap between mathematically correct kinematics and perceptually realistic 3D animation. Table~\ref{tab:abl_table} reports the quantitative results for this setting, showing strong text alignment with a minor increase in distortion introduced by secondary motion effects.

\subsection{Limitation and future work}
Despite its effectiveness, the proposed diffusion-guided, physics-aware framework has certain limitations that suggest promising directions for future research. First, our method relies on pretrained text-to-video diffusion models as semantic motion priors. While this enables motion synthesis, the quality and diversity of generated motion are inherently bounded by the capabilities and biases of the underlying text-to-video prior model. Complex or highly specialized motions that are underrepresented in the training data may not be accurately captured, leading to suboptimal or ambiguous motion guidance. Additionally, the diffusion-guided optimization process is computationally expensive, requiring repeated rendering and gradient-based updates across long animation sequences, limiting scalability to high-resolution meshes or real-time applications. Second, although we incorporate physics-inspired constraints to stabilize optimization, our formulation does not explicitly model full physical dynamics such as forces, mass, or contact interactions. As a result, certain motion artifacts such as subtle ground sliding or inaccurate contact timing may still occur in challenging scenarios. Furthermore, the skeleton extraction and rigging process depends on the quality of the reconstructed 3D mesh; errors in sketch-to-3D reconstruction or skeleton inference can propagate into the animation stage. Finally, our current framework focuses on single-character animation and does not explicitly handle interactions between multiple animated entities or dynamic environments.

Future work can explore several directions to address these limitations. Recent closed-source text-to-video (T2V) diffusion models can further improve motion generation capabilities. Furthermore, incorporating lightweight physics simulators or learned dynamics models could improve contact handling and physical realism while maintaining differentiability. Motion efficiency may be enhanced by distilling diffusion guidance into motion priors or adopting latent-space optimization strategies to reduce computational overhead. Extending the framework to multi-character interactions and environment-aware motion generation represents another promising direction. Finally, integrating richer conditioning signals, such as user-provided keyframes or sketch-based motion hints, could further enhance controllability and expand the applicability of the proposed approach.

\section{Conclusion}
We introduce a diffusion-guided, physics-aware framework for skeleton-based 3D animation. Our method works directly on explicit skeletal motion parameters, including joint rotations and root translations, and integrates forward kinematics, linear blend skinning, and differentiable rendering into a unified optimization framework. By integrating motion-aware score distillation guidance from a text-to-video diffusion model, the proposed approach injects high-level semantic motion cues while preserving the interpretability and controllability of skeleton-based representations. To ensure stable optimization and physically plausible animation, we incorporate a set of physics-inspired constraints that enforce temporal smoothness, joint motion limits, structural consistency, and contact-aware motion. We integrate the SpringMass simulator to model secondary motion effects, making the animation more realistic. This combination enables our framework to generate realistic, temporally coherent motion while avoiding common artifacts, such as jitter, distortion, and kinematic violations. Importantly, our approach does not require paired motion data or task-specific retraining and generalizes across diverse articulated structures, including bipeds, quadrupeds, and non-living objects. Together, these contributions establish a flexible and principled framework for controllable, semantically meaningful motion transfer driven by diffusion priors.


\bibliography{main.bib}
\bibliographystyle{ieeetr}

\end{document}